\documentclass{article}
\usepackage[utf8]{inputenc}
\usepackage{authblk}
\usepackage{setspace}
\usepackage[margin=1.25in]{geometry}
\usepackage{graphicx}
\graphicspath{ {./figures/} }
\usepackage{subcaption}
\usepackage{amsmath}
\usepackage{amssymb}
\usepackage{lineno}
\usepackage{hyperref}
\usepackage{float}

\usepackage[style=ieee, 
citestyle=numeric-comp,
sorting=none]{biblatex}
\title{Multi-style Neural Radiance Field with AdaIN}

\author[1*$\dagger$]{
Yu-Wen, Pao
}
\author[1*$\ddagger$]{
An-Jie, Li
}

\affil[1]{Computer Science and Information Engineer, National Taiwan University.}
\affil[$\dagger$]{b09902016@csie.ntu.edu.tw}
\affil[$\ddagger$]{b09902017@csie.ntu.edu.tw}
\affil[*]{These authors contributed equally to this work.}

\date{}

\begin{document}

\maketitle

\begin{abstract}
In this work, we propose a novel pipeline that combines AdaIN and NeRF for the task of stylized Novel View Synthesis\footnote{The implementation is available at \href{https://github.com/paoyw/Stylized-NeRF-with-AdaIN}{https://github.com/paoyw/Stylized-NeRF-with-AdaIN}}. Compared to previous works, we make the following contributions: 1) We simplify the pipeline. 2) We extend the capabilities of model to handle the multi-style task. 3) We modify the model architecture to perform well on styles with strong brush strokes. 4) We implement style interpolation on the multi-style model, allowing us to control the style between any two styles and the style intensity between the stylized output and the original scene, providing better control over the stylization strength.

\end{abstract}


\section{Introduction}
Stylized Novel View Synthesis is a technique that takes real-world scene photographs as input and generates stylized images of the same scene from different viewpoints. This technology enables various applications in computer graphics by creating virtual scenes with artistic styles based on real-world environments.

Previous works have explored the combination of AdaIN and NeRF techniques for stylized Novel View Synthesis. However, we found that previous architectures were complex, and each model could only support a single style. We aimed to design a simpler approach that achieves comparable results to previous works while allowing the model to support multiple styles.

To achieve our objectives, we incorporate the style feature extracted by AdaIN and the voxels information from NeRF into the final layers of the NeRF model (which are Multilayer Perceptrons (MLP) used to predict color and density of the voxel) . This integration allows us to stylize the original scene voxel-wise and ultimately achieve the stylized rendering of the entire scene.

Through extensive experimentation, our proposed pipeline effectively stylizes scenes and performs well when switching between different styles or training with multiple styles. Additionally, we implemented style interpolation functionality on our multi-style model, allowing us to blend two different styles or control the degree of stylization through interpolation between the stylized scene and the original scene.

\section{Related Works}
\subsection{Image Style Transfer}
In recent years, image style transfer has been a popular research topic in computer vision and image processing. One line of work focuses on deep learning-based methods. These approaches aim to learn representations that capture the style and content of input images separately, enabling the style transfer between different contents images. Gatys et al.\cite{gatys2015neural} introduced a neural network-based method, which utilized deep convolution neural networks to separate the representations of the content and style images and then recombine them to create visually appealing stylized outputs. Huang et al.\cite{huang2017arbitrary} introduced AdaIN which enables the style transfer process by dynamically adjusting the mean and variance of features in a deep network, effectively transferring the style statistics from a style image to a content image. Zhu et al.\cite{zhu2020unpaired} proposed an unsupervised image-to-image translation framework, which leverages the power of generative adversarial networks to learn mapping functions between two domains.
\subsection{Novel View Synthesis}
Novel view synthesis aims to generate visually consistent and plausible images from unseen viewpoints using limited input information. In recent years, due to the flourishing of deep learning, neural networks have brought new light to this field. NeRF\cite{mildenhall2021nerf} uses MLPs to learn the mapping function between positions and view directions to color and density.

\subsection{Style-transfer with NeRF}
Chiang et al.\cite{chiang2022stylizing} is one of the pioneers to combine style transfer and NeRF. Their framework consists of two networks, a NeRF to find the implicit representation of the 3D scene and a hypernetwork to transfer the style. What's more, they use two-stage training and a patch sub-
sampling approach procedure to fix the problem of memory burden and training difficulties.

Huang et al.\cite{huang2022stylizednerf} combined NeRF and AdaIN using mutual learning. After training the vanilla NeRF and AdaIN, it mutually trains the style module, its inputs, which are learnable latent code, and the decoder of AdaIN with a combination of four loss terms.

Zhang et al.\cite{zhang2022arf} introduced a nearest neighbor-based loss to capture style details while maintaining multi-view consistency. What's more, they proposed the deferred back-propagation method to make optimization of memory-intensive NeRF using style losses defined on full-resolution rendered images possible.

However, most of the methods mentioned above ignore that density in 3D scenes can be an important role in some styles. Also, they cannot transfer a single scene to multiple styles using only one model.

\section{Method}
We propose a simple framework combining AdaIN and NeRF to achieve multi-style transfer for novel view synthesis. To be more specific, we use a three-stage training procedure. First, the 2D stylization training stage learns the generation of stylized images by AdaIN. Secondly, the geometric training stage learns the mapping function of position and view direction to opacity and radiance color of a 3D scene by NeRF. Finally, the 3D scene stylization training stage uses the camera parameters and its corresponding image after stylization as inputs and takes the MLPs with part of pre-trained NeRF from the previous stage as the network to learn the style transfer of a 3D scene.

\subsection{2D Stylization Network}
For 2D stylization network, we use AdaIN\cite{huang2017arbitrary} as our method. First, AdaIN uses pre-trained VGG\cite{simonyan2015deep} as encoder to extract the features of the content and style images. Secondly, adaptive instance normalize layer aligns the channel-wised means and variance of content image to those of style image. Finally, a CNN-based decoder will decode the transformed features to stylized images. The loss function to train the decoder $\mathcal{L}=\mathcal{L_c}+\lambda\mathcal{L_s}$, where $\mathcal{L_c}$ is the content loss and $\mathcal{L_s}$ is the style loss. The content loss $\mathcal{L_c}$ aligns the feature to reconstruct stylized images and the feature of the stylized images. Style loss $L_s$ aligns the means and variances of the features of style images and stylized images.
We train an AdaIN using MS-COCO\cite{lin2015microsoft} as our content images and 100 images mostly from \href{https://www.wikiart.org/}{WikiArt} as our style images.

\subsection{NeRF}
NeRF\cite{mildenhall2021nerf} uses MLPs to model the randiance color and opacity of a 3D scene. The MLPs take 3D position $\textbf{x}\in\mathbb{R}^3$ and viewing direction $\mathbf{b}\in\mathbb{R}^2$ as inputs. The outputs of MLPs are opacity $\sigma(\mathbf{x})\in\mathbb{R}$ and radiance color $c(\mathbf{x}, \mathbf{d})\mathbb{R}^3$. The following integral tells us how to obtain the single pixel on the image from the radiance color and opacity along a ray.
$$
C(\mathbf{r})=\int_{t=0}^{\infty}\sigma(\mathbf{o}+t\mathbf{d}, \mathbf{d})\exp{(-\int_{s=0}^t\sigma(\mathbf{o}+s\mathbf{d})ds)}
$$

To estimate this continuous integral, they uses quadrature. More specific, they use a two-stage sampling strategy, a stratified sampling, which sample one sample in evenly-spaced bins, and hierarchical volume sampling, which tends to samples more points biased towards the parts with high density.

To improve the performance, NeRF also uses positional encoding $\gamma(\cdot)$ to map $\mathbf{x}$ and $\textbf{d}$ to Fourier features.
$$
\gamma(x)=(\sin{(x)}, \cos{(x)},\dots, \sin{(2^{L-1}x)},\cos{(2^{L-1}x)})
$$

We take the PyTorch implementation of NeRF from \cite{lin2020nerfpytorch}.

\subsection{Multi-style NeRF}
\begin{figure}[H]
    \centering
    \includegraphics[scale=0.4]{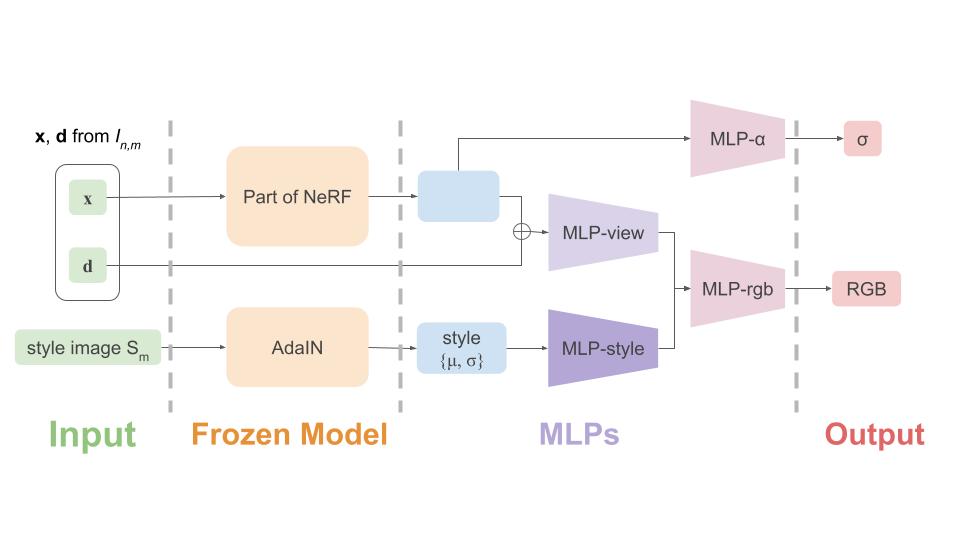}
    \caption{Multi-style NeRF}
    \label{fig:multi_nerf}
\end{figure}

Given a set of N images $\{\mathcal{I}_n\}_{n=1}^N$ and their corresponding camera poses $\{(R_n, t_n)\}_{n=1}^N$. We train the vanilla NeRF using $\{\mathcal{I}_n\}_{n=1}^N$ and $\{(R_n, t_n)\}_{n=1}^N$. Then, we stylize the images, $\{I_n\}_{n=1}^N$, to stylized images, $\{I_{n,m}\}_{n=1,m=1}^{N,M}$, with $M$ style images $\{S_m\}_{m=1}^M$ with pre-trained AdaIN. We take the stylized images $\{I_{n,m}\}_{n=1,m=1}^{N,M}$ and their coresponding camera poses $\{(R_{n,m}=R_n, t_{n,m}=t_n)\}_{n=1,m=1}^{N,M}$ as the new dataset.

For model architecture, we take the top few layers of NeRF\cite{mildenhall2021nerf}, which have been trained with $\{\mathcal{I}_n\}_{n=1}^N$ and $\{(R_n, t_n)\}_{n=1}^N$, and frozen their weight. For density, the output feature of the frozen layers will be passed to a MLP $MLP_\alpha$ to predict the density. For color prediction, we will first extract the feature of the style images using a frozen VGG-based model like \cite{huang2017arbitrary}. As mentioned in \cite{7780634,li2016combining,li2017demystifying,huang2017arbitrary}, feature statistics can carry the style information of an image. Hence, we use another MLP $MLP_{style}$ to extract the feature of the feature statistics extracted from the VGG-based model. In original NeRF\cite{mildenhall2021nerf}, the color prediction will take the view direction $\mathbf{d}$ as one of the inputs. We follow the same concept by concatenating the output of the frozen NeRF with view direction $\mathbf{d}$, passing it to a MLP $MLP_{view}$. Then, we concatenate the output of $MLP_{style}$ and the output of $MLP_{view}$ and pass it to another MLP $MLP_{rgb}$ to predict the color of a voxel.

To render a pixel from an image, we use the integral the same as the original NeRF\cite{mildenhall2021nerf} with the same sampling strategy.
$$
\hat{C}(\mathbf{r}, I_{style})=\sum_{i=1}^N T_i(1-exp(-\sigma_i\delta_i))\mathbf{c_i}\textit{, where }T_i=exp(-\sum_{j=1}^{i-1}\sigma_j\delta_j) 
$$

For loss function, we use the same function as the original NeRF\cite{mildenhall2021nerf} and no need for explicit loss function for style transfer as other 3D style transfer method\cite{chiang2022stylizing,huang2022stylizednerf,zhang2022arf}.

$$
\mathcal{L}=\sum_{i=1}^N\sum_{j=1}^M\sum_{\mathbf{r}\in\mathcal{R}_i}\left[\parallel\hat{C}(\mathbf{r}, I_{i,j})-C(\mathbf{r}, I_{i,j})\parallel\right]
$$

For the other hyper-parameters, we follow the same as the original NeRF\cite{mildenhall2021nerf}.
\subsubsection*{Density Aware}
\begin{figure}[H]
    \centering
    \includegraphics[scale=0.4]{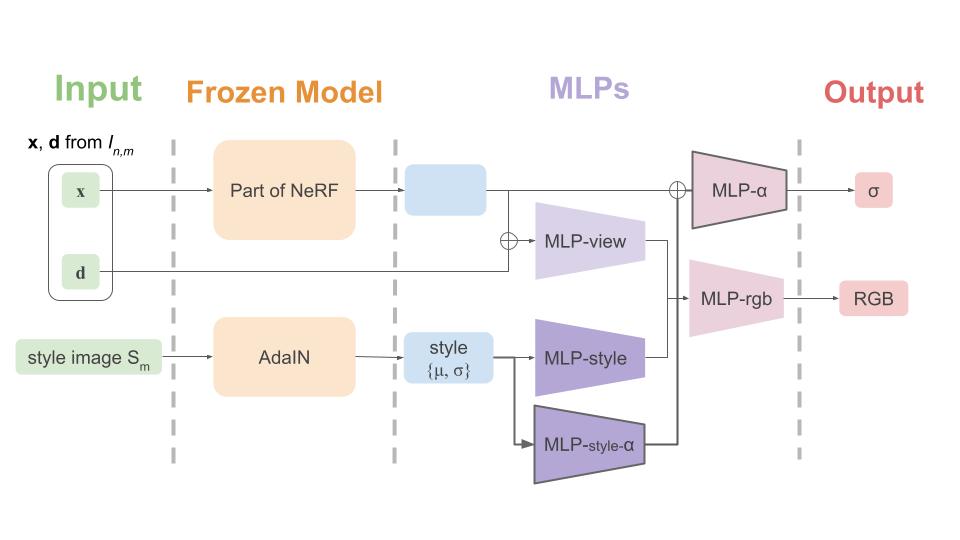}
    \caption{Multi-style NeRF with Density Aware}
    \label{fig:multi_nerf_density}
\end{figure}

To make the prediction of density vary between different style images, we modify the framework to \ref{fig:multi_nerf_density}. First, we add a new MLP to transform the style features captured by AdaIN to the features for the density prediction. Then, we concatenate the output of this MLP with the output of the pre-trained NeRF. Finally, we feed the concatenated features to a MLP to predict the density of a voxel. The remain training procedure is the same as the one mentioned above.

\subsubsection*{Style Control and Style Interpolation}
To do the style interpolation, we can easily interpolate the feature statistics of the style image ${\mathcal{S}_m}_{m=1}^M$.
To control the balance of the content and style, we can train the multi-style NeRF by taking the original content image $\mathcal{I}_n$ as another style. Then, we directly interpolate the feature statistics of the original image and the style image.

\section{Results}
We invite readers to watch our videos from our \href{https://github.com/vppyw/Stylized-NeRF-with-AdaIN}{Github}  for better assessment of 3D stylization
quality.

\section{Conclusion}
We present Multi-style NeRF, which is a novel pipeline for stylized novel view synthesis that combines AdaIN and NeRF techniques. Our contributions include simplifying the pipeline of the previous stylization of novel view synthesis, extending its capabilities to handle multiple styles, modifying the model architecture to handle styles with multiple styles, and implementing style interpolation for better control over stylization strength.
\printbibliography
\end{document}